\documentclass[letterpaper, 10 pt, conference]{ieeeconf}  

\usepackage[ruled,linesnumbered]{algorithm2e}
\usepackage{algorithmic}
\usepackage{graphics} 
\usepackage{subfigure}
\usepackage{epsfig} 
\usepackage{mathptmx} 
\usepackage{times} 
\usepackage[tbtags]{amsmath} 
\usepackage{amssymb}  
\usepackage{bm}
\usepackage{multirow}
\usepackage{booktabs}
\usepackage{array}
\usepackage{tabularx,booktabs}
\usepackage{overpic}
\usepackage{url}
\usepackage[bookmarks=false]{hyperref}
\usepackage{rotating}
\usepackage{mathtools}
\usepackage{authblk}
\usepackage{cite}
\IEEEoverridecommandlockouts                              

\title{\LARGE \bf
Robot Navigation with Map-Based  Deep Reinforcement Learning
}

\author{Guangda Chen, Lifan Pan, Yu'an Chen, Pei Xu, Zhiqiang Wang, Peichen Wu, \\ Jianmin Ji$^*$ and Xiaoping Chen
\thanks{
        This work is partially supported by the 2030 National Key AI Program of China 2018AAA0100500, 
        the National Natural Science Foundation of China (No. 61573386), and Guangdong Province Science and Technology Plan Projects (No. 2017B010110011).
}
\thanks{The authors are with the School of Computer Science and Technology, University of Science and Technology of China, Hefei, Anhui, 230026, P.R.China. 
{\tt\small \{cgdsss, lifanpan, an11099, xp816, tt1248163264, wpc16\}@mail.ustc.edu.cn, 
\{jianmin, xpchen\}@ustc.edu.cn.}}
\thanks{ * Corresponding author.}

}

\begin{document}

\maketitle
\thispagestyle{empty}
\pagestyle{empty}

\begin{abstract}
        This paper proposes an end-to-end deep reinforcement learning approach for mobile robot navigation 
        with dynamic obstacles avoidance. Using experience collected in a simulation environment, a convolutional 
        neural network (CNN) is trained to predict proper steering actions of a robot from its egocentric local occupancy 
        maps, which accommodate various sensors and fusion algorithms. The trained neural network is then transferred and 
        executed on a real-world mobile robot to guide its local path planning. The new approach is evaluated both qualitatively 
        and quantitatively in simulation and real-world robot experiments. The results show that the map-based end-to-end navigation 
        model is easy to be deployed to a robotic platform, robust to sensor noise and outperforms other existing DRL-based models 
        in many indicators.
\end{abstract}

\section{Introduction}
\label{sec:intro}
One of the major challenges for robot navigation is to develop a safe and robust collision avoidance policy to navigate 
from the starting position to desired goal without running into obstacles and pedestrians in unknown cluttered environments. 
Although numerous methods have been proposed \cite{mohanan2018survey}, conventional methods are often built upon a set of assumptions that are likely 
not to be satisfied in practice 
\cite{zhang2017dynamic}, and may impose intensive computational demand \cite{zhou2017fast}. In addition, conventional algorithms 
normally involve a number of parameters that need to be tuned manually \cite{rosmann2017integrated} rather than being able to learn from 
past experience automatically \cite{kahn2018self}. It is difficult for these approachs to generalize well to unanticipated scenarios. 

Recently, several supervised and self-supervised deep learning approaches have been applied to robot navigation. 
Giusti et al. \cite{giusti2015machine} used a Deep Neural Network  to classify the images to determine which action will
keep the quadrotor on the trail. 
Lei et al. \cite{tai2016deep} showed the effectiveness of a
hierarchical structure that fuses a convolutional neural network (CNN) with a decision process, which is a highly 
compact network structure that takes raw depth images as input, and generates control commands as network output. 
Pfeiffer et al. \cite{pfeiffer2017perception} presented a model that is able to learn the complex mapping from 
raw 2D-laser range findings and a target position to the required steering commands for the robot.
However, there are some limitations that prevent these approaches from being widely applied in a real robotic setting. 
For instance, a massive manually labeled dataset is required for 
the training of the supervised learning approaches. Although this can be migrated to an extent by resorting to self-supervised learning methods, their 
performance is largely bounded by the strategy generating training labels.

On the other hands, deep reinforcement learning (DRL) methods have achieved remarkable success in many challenging tasks \cite{silver2018general, levine2018learning, vinyals2019grandmaster}.
Different from previous supervised learning methods, DRL based approaches learn from a large number of trials and 
corresponding rewards instead of labeled data. 
In order to learn a sophisticated control policy with reinforcement learning, robots need to 
interact with the environment for a long period to accumulate knowledge about the consequences of different actions. 
Collecting such interaction data in real world is expensive, time consuming, and sometimes infeasible 
due to safety issues {\cite{xie2018learning}}. 
For instance, Kahn et al. \cite{kahn2018self} proposed a generalized computation graph that subsumes value-based
model-free methods and model-based methods, and then instantiated this graph to form a navigation model that
learns from raw images and is sample efficient. However, It takes hours of destructive self-supervised training to
navigate only dozens of meters without collision through a indoor environment.
Because of the excessive number of trials required to learn a good policy, training in a simulator is more 
suitable than experiences derived from the real world. Then we can exploit the close correspondence between 
a simulator and the real world, to transfer the learned policy.

According to the difference of input data, the existing reinforcement learning-based robot motion planning methods 
can be roughly divided into two categories: agent-level inputs and sensor-level inputs. And there are different 
transferability to real world between different input data. As representatives of agent-level methods, 
Chen et al. \cite{chen2017socially} train an agent-level collision avoidance policy using DRL, which maps 
an agent's own state and its neighbors' states to collision-free action. However, it demands perfect sensing. 
Obviously, this complex pipeline 
not only requires expensive online computation but makes the whole system less robust to the perception uncertainty.

As for sensor-level inputs, the types of sensor data used in DRL-based navigation mainly include 2D laser range inputs, 
depth images and color images. The network proposed in \cite{tai2017virtual} outputs control commands based on ten-dimensional
laser range inputs and is trained using an asynchronous DRL algorithm. Similarly, the models introduced in \cite{long2018towards} 
and \cite{xie2018learning} also derive the steering commands from laser range sensors. 
2D laser-based methods are competitive in terms of the transferability to real world because
of the smaller discrepancies between their simulated and real domains. However, 2D sensing data is not enough to describe 
complex 3D scenarios. On the contrary, vision sensors can provide 3D sensing informations. But
RGB images suffer from the significant
deviation between real-world situations and the simulation environments during training, which leads
to quite limited generalization across situations. Compared to RGB images, the depth inputs in
simulations exhibit much better visual fidelity due to the textureless nature and, as a result, greatly
alleviate the burden of transferring the trained model to real deployments \cite{wu2018learn}. 
Based on depth images,   
Zhang et al. \cite{zhang2017deep} proposed to use successor features to achieve efficient knowledge transfer across tasks in 
depth-based navigation. Currently, all existing sensor-level works rely on specific sensor types and configurations. 
While in complex environments, most robots are equipped with different sensors to navigate autonomously and 
safely \cite{chen2018accurate}.

In this paper,  we propose an end-to-end model-free deep reinforcement learning 
algorithm to improve the performance of autonomous decision making in complex environments, 
which directly maps local probabilistic costmaps to an agent's steering commands in terms of target position and 
robot velocity. Compared to previous 
work on DRL-based obstacle avoidance, our motion planner is based on probabilistic costmaps to represent environment and 
target position, which enables the learned collision avoidance policy to handle different types of sensor input efficiently, 
such as the multi-sensor information from 2D/3D range scan finders or RGB-D cameras \cite{liu2018map}.
And our trained CNN-based policy is easily transferred and executed on a real-world mobile robot to guide its local path planning and 
robust to sensor noise.  
We evaluate our DRL agents both in 
simulation and on-robot qualitatively and quantitatively. Our results show the improvement in multiple indicators over the DRL-based 
obstacle avoidance policy.

Our main contributions are summarized as follow: 
\begin{itemize}
        \item Formulate the obstacle avoidance for mobile robots as an DRL problem based on a generated costmap, which 
        can handle multi-sensor fusion and is robust to sensor noise.
        \item Integrate curriculum learning technique to enhance the performance of dueling double DQN 
        with prioritized experience reply.
        \item Contract a variety of real-world experiments to reveal the high adaptability of our model to 
        transfer to different sensor configurations and environments.
\end{itemize}

The rest of this paper is organized as follows. The structure of the DRL-based navigation system is presented in 
Section \ref{sec: ss}. The deep reinforcement learning algorithm for obstacle avoidance based on egocentric local occupancy maps 
is described in Section \ref{sec:approach}. Section \ref{exp} presents experimental results, followed by conclusions in 
Section \ref{conclusion}.

\section{System Structure}
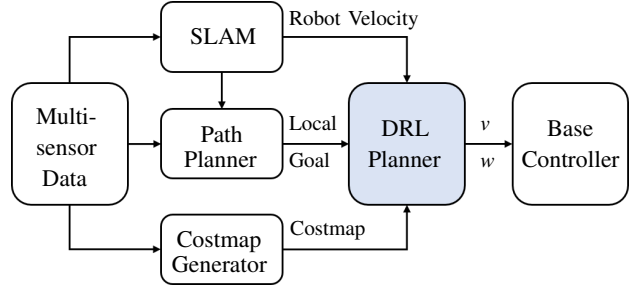
\begin{figure}
        \centering
        \begin{overpic}[width=0.95\linewidth ]{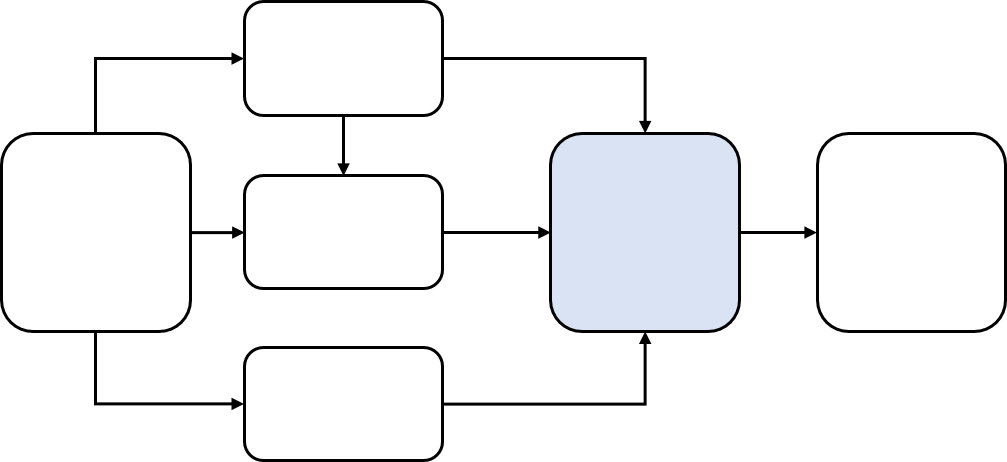}
                \put(4,26){\fontsize{9pt}{\baselineskip}\selectfont{Multi-}}
                \put(4,21){\fontsize{9pt}{\baselineskip}\selectfont sensor}
                \put(5,16){\fontsize{9pt}{\baselineskip}\selectfont Data}
                \put(29,39){\fontsize{9pt}{\baselineskip}\selectfont SLAM}
                \put(30.7,23){\fontsize{9pt}{\baselineskip}\selectfont Path}
                \put(28,19){\fontsize{9pt}{\baselineskip}\selectfont Planner}
                \put(26.8,6){\fontsize{9pt}{\baselineskip}\selectfont Costmap}
                \put(26,2){\fontsize{9pt}{\baselineskip}\selectfont Generator}
                \put(45,42){\footnotesize Robot Velocity}
                \put(45,25){\footnotesize Local}
                \put(45,19){\footnotesize Goal}
                \put(45.2,8){\footnotesize Costmap}
                \put(76,25){\footnotesize $v$}
                \put(76,19){\footnotesize $w$}
                \put(60,24){\fontsize{9pt}{\baselineskip}\selectfont DRL}
                \put(57.8,19){\fontsize{9pt}{\baselineskip}\selectfont Planner}
                \put(86.8,24){\fontsize{9pt}{\baselineskip}\selectfont Base}
                \put(82.7,19){\fontsize{9pt}{\baselineskip}\selectfont Controller}
        \end{overpic}
        \caption{Block diagram of the DRL-based navigation system for autonomous robots. }
        \label{figflow}
\end{figure}
\label{sec: ss}
The proposed DRL-based mobile robot navigation system consists of six modules. As shown in Fig. \ref{figflow}, 
simultaneous localization and mapping (SLAM) module establishes an environment map based on sensor data, and 
can estimate the position and velocity of the robot in 
the map simultaneously. When a target position is received,  the path planner module generates a path or a series of local goal points
from the current position to the target position. In order to cope with the dynamic and complex environments, 
a safe and robust collision avoidance policy in unknown cluttered environments is required. 
In addition to the local goal points from the path planning module and  
robot velocity provided by the positioning module, our local planner also needs the input of the surrounding environment information,
which is the egocentric occupancy map produced by the costmap generator module based on multi-sensor data. Generally speaking, 
Our DRL-based local planner takes the velocity information generated by the SLAM module, the local goal generated by the global path planner and the 
cost maps from the generator which can fuse multi-sensor information, and outputs the linear velocity $v$ and angular velocity $w$ of 
the robot. Finally, the output speed command is executed by the base controller module which depends on the specific kinematics of the robot.

\section{Approach}
\label{sec:approach}
We begin this section by introducing the problem formulation of the local obstacle avoidance.
Next, we describe the key ingredients of our reinforcement learning algorithm and the 
details about the network architecture of the collision avoidance policy.
\subsection{Problem Formulation}
\label{sec:problem}
At each timestamp $t$, given a frame sensing data $s_t$, 
a local goal position $g_t$ in the robot coordinate system and the 
robot linear velocity $v_t$, angular velocity ${\omega}_t$, the proposed 
local obstacle avoidance policy provides an action command $a_{t}$ as follows:
\begin{equation}
        \textbf{M}_t = f_{\lambda}(s_t)
\end{equation}
\begin{equation}
        a_{t} = \pi_{\theta}(\textbf{M}_t, g_t, v_t, {\omega}_t)
\end{equation}
where $\textbf{M}_t$ is a local cost map describing the obstacle avoidance task, $\lambda$ and $\theta$ are model parameters.
Specifically, the cost map $\textbf{M}_t$ is constructed as an aggregate of robot configurations and 
the obstacle penalty, which will be explained below.

Hence, the robot collision avoidance problem can be simply formulated as a sequential decision making problem. The 
sequential decisions consisting of observations $\textbf{o}_t \sim [\textbf{M}_t, g_t, v_t, {\omega}_t]$ and actions (velocities)
$a_t \sim [v_{t+1}, {\omega}_{t+1}]$ ($t=0:t_g$) can be considered as a trajectory $l$ from its start position $\textbf{p}_0$ 
to its desired goal $\textbf{p}_g$, where $t_g$ is the traveled time. Our goal is to minimize the expectation of the 
arrival time and take into account that robots do not collide, which is defined as: 
\begin{equation}
        \begin{aligned}
                \underset{\pi _\theta }{\textrm{argmin}} \mathbb{E}[t_g | & a_t = \pi _\theta(\textbf{o}_t),\\ 
                &\textbf{p}_t = \textbf{p}_{t-1} + a_t \cdot \Delta  t, \\
                &\forall k\in [1, N] : \left \| \textbf{p}_t - (\textbf{p}_{obs})_k \right\| > R]   
        \end{aligned}
\end{equation}
where $\textbf{p}_{obs}$ is the position of obstacle and $R$ is the robot radius.
\subsection{Dueling DDQN with prioritized experience reply}
Markov Decision Processes (MDPs) provide a mathematical framework to model stochastic planning and decision-making problems 
under uncertainty. An MDP is a tuple $M = (\textit{S},\textit{A},\textit{P},\textit{R},\gamma)$, where $\textit{S}$ indicates 
the state space, $\textit{A}$ is the action space, $\textit{P}$ indicates the transition function which describes the 
probability distribution over states if an action $\textit{a}$ is taken in the current state $\textit{s}$, $\textit{R}$ represents 
the reward function which illustrates the immediate state-action reward signal, and ${\gamma} \in [0, 1]$ is the 
discount factor. In an MDP problem, a policy ${\pi}(a|s)$ specifies the probability of mapping state $s$ to action $a$. The 
superiority of a policy $\pi$ can be assessed by the action-value function (Q-value) defined as: 
\begin{equation}
        Q^{\pi}(s,a) = {\mathbb{E}^{\pi}}{[\sum_{t=0}^{\infty }{\gamma ^t}{R(s_{t},a_t)}|s_0=s,a_0=a]}
\end{equation}

Therefore, the action-value function is the expectation of discounted sums of rewards, given that ation $a$ is taken in 
state $s$ and policy $\pi$ is executed afterwards. The objective of the agent is to maximize the expected cumulative 
future reward, and this can be achieved by adopting the Q-learning algorithm which approximates the optimal action-value 
function iteratively using the Bellman equation: 
\begin{equation}
        Q^{*}(s_t,a_t) = R(s_t,a_t) + \gamma {\underset{a_{t+1}}{\textrm{max}}}Q(s_{t+1}, a_{t+1})
\end{equation}

Combined with deep neural networks, DQN enables reinforcement learning to cope with complex high-dimensional problems. Generally, 
DQN maintains two deep neural networks, including an online network with parameters $\theta$ and a separate target network 
with parameters $\theta{'}$. The parameters of the online network are updated constantly by minimizing the loss function 
$(y_t - Q(s_t, a_t; \theta'))^2$, where $y_t$ can be calculated as follows: 
\begin{equation}
        y_t = \left\{\begin{matrix*}[l]
                r_t & \textrm{{if episode ends}} \\  
                r_t + \gamma \underset{a_{t+1}}{\textrm{max}}Q(s_{t+1}, a_{t+1}; \theta') & \textrm{otherwise}
                \end{matrix*}\right.
\label{dqn}
\end{equation}
And the parameters of the target network are fixed for generating Temporal-Difference (TD) targets and synchronized regularly 
with those of the online network.

Conventional Q-learning is affected by an overestimation bias, due to the maximization step in
Equation (\ref{dqn}), which would harm the learning process.  Double Q-learning \cite{van2016deep}, addresses this overestimation by decoupling, 
in the maximization performed for the bootstrap target, the selection of the action from its evaluation. 
Therefore, if episode not ends, the $y_t$  in the above formula is rewritten as follows:
\begin{equation}
        y_t = 
                r_t + \gamma Q(s_{t+1}, \underset{a_{t+1}}{\textrm{argmax}}Q(s_{t+1}, a_{t+1}; {\theta}); {\theta}')
\label{ddqn}
\end{equation}

In this work, dueling networks \cite{pmlr-v48-wangf16} and prioritized replay \cite{Schaul2016} are also deployed for more reliable estimation of Q-values and 
sampling efficiency of replay buffer respectively. In the following, we describe the details of the observation space, 
the action space, the reward function and the network architecture.

\subsubsection{Observation space}
\begin{figure}
        \centering
        \subfigure[] {\includegraphics[width = 0.48\linewidth]{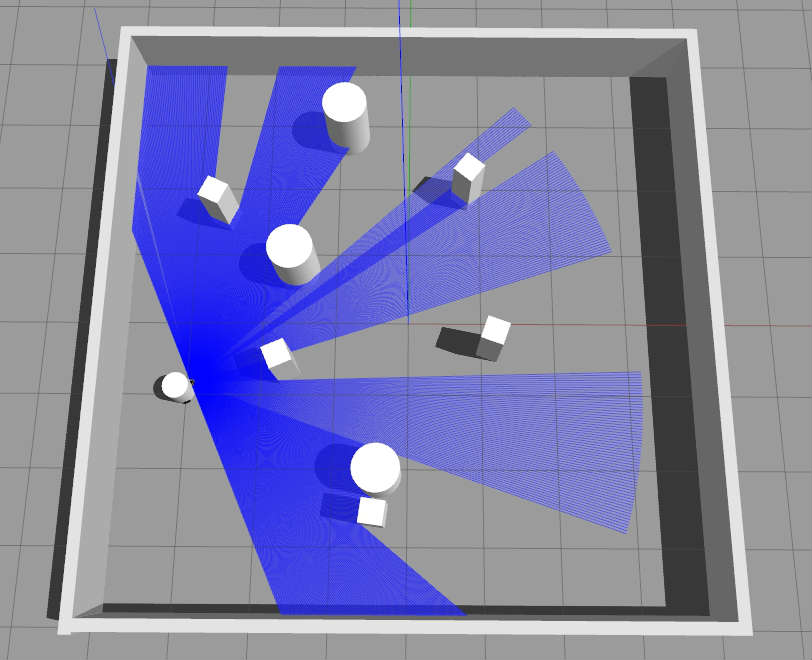}\label{fig:scenarios:gazebo}}
        \subfigure[] {\includegraphics[width = 0.48\linewidth]{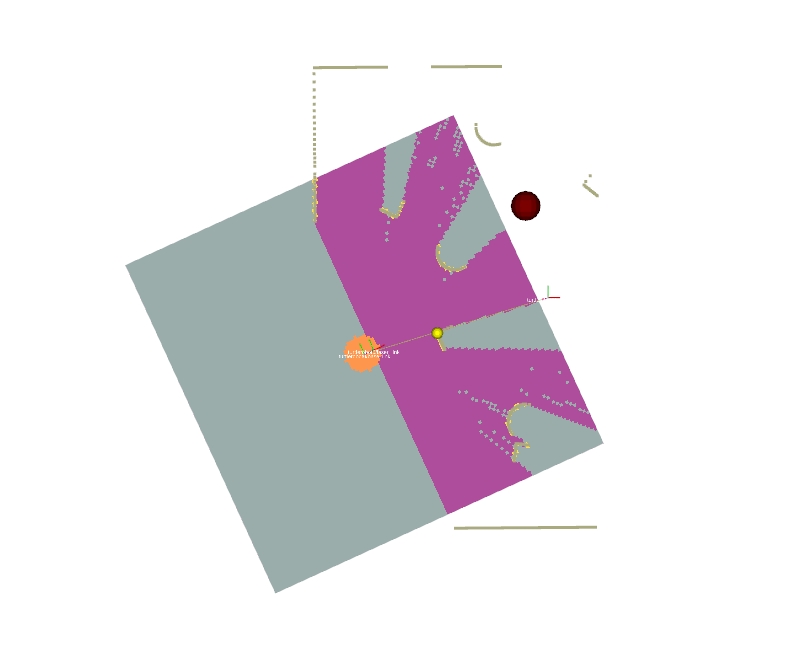}\label{fig:scenarios:rviz}}
        \caption{Gazebo training environments (left) and corresponding occupancy map displayed by Rviz (right).}
        \label{fig:scenarios}
\end{figure}
As mentioned in Section \ref{sec:problem}, the observation $\textbf{o}_t$ consists of the generated costmaps 
$\textbf{M}_{t}$, the relative goal position $\textbf{g}_{t}$ and the robot's current velocity 
$\textbf{v}_{t}$. Specifically, $\textbf{M}_{t}$ represents the cost map images generated 
from a 180-degree laser scanner or other sensors. The relative goal position $\textbf{g}_{t}$ is a 2D vector 
representing the goal coordinate with respect to the robot's current position. The observation $\textbf{v}_{t}$ 
includes the current transitional and rotational velocity of the differential-driven robot.

We use layered costmaps \cite{lu2014layered} to represent environmental information perceived by multi-sensors. 
Then though the map generater module, we get the state maps $\textbf{M}_t$ by drawing 
the robot configuration (shape) into the layered costmaps. Fig. \ref{fig:scenarios:rviz} shows an example 
of generated occupancy maps.
\subsubsection{Action space}
The action space is a set of permissible velocities in discrete space. The action of differential
robot includes the translational and rotational velocity, i.e.
$\textbf{a}_t = [v_{t}, w_{t}]$. In this work, considering the real robot’s
kinematics and the real world applications, we set the range
of the translational velocity $v \in [0.0, 0.2, 0.4, 0.6]$ and the rotational
velocity in $w \in [−0.9, -0.6, -0.3, 0.0, 0.3, 0.6, 0.9]$. Note that backward moving (i.e.
$v < 0.0$) is not allowed since the laser range finder can not
cover the back area of the robot.
\subsubsection{Reward}

\begin{figure*}
        \centering
        \begin{overpic}[width=0.95\linewidth]{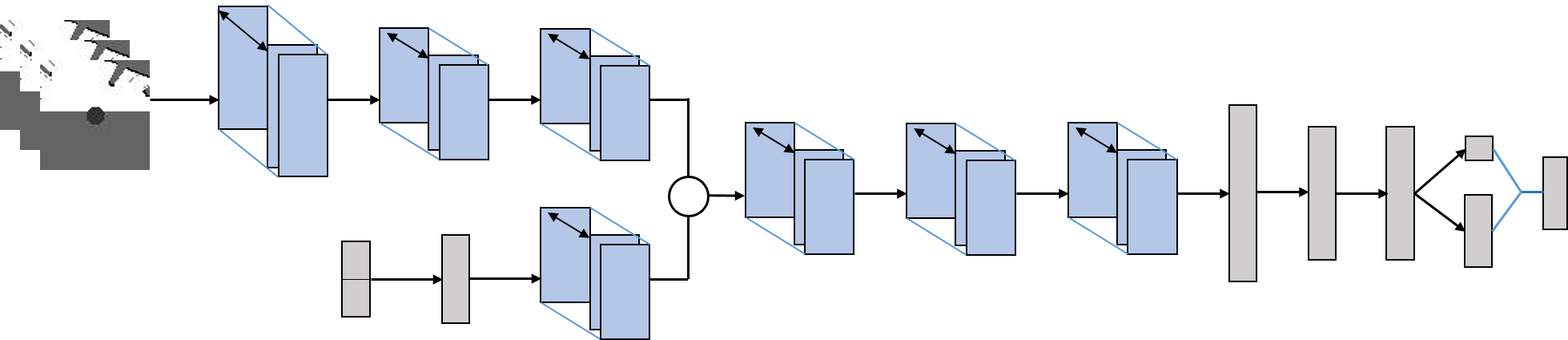}
        \put(10,16){\begin{turn}{90} \footnotesize stride \scriptsize 4\end{turn}}
        \put(11.5,16){\begin{turn}{90} \scriptsize {8x8} \footnotesize{conv}\end{turn}}
        \put(10.5,14){\tiny {ReLU}}
        \put(15.5,20){\scriptsize 32 filters}

        \put(21.4,16){\begin{turn}{90} \footnotesize stride \scriptsize 2\end{turn}}
        \put(22.9,16){\begin{turn}{90} \scriptsize {4x4} \footnotesize{conv}\end{turn}}
        \put(21.2,14){\tiny {ReLU}}
        \put(26.5,18.8){\scriptsize 64 filters}

        \put(31.5,16){\begin{turn}{90} \footnotesize stride \scriptsize 1\end{turn}}
        \put(33,16){\begin{turn}{90} \scriptsize {3x3} \footnotesize{conv}\end{turn}}
        \put(31.5,14){\tiny {ReLU}}
        \put(36.5,18.8){\scriptsize 64 filters}

        \put(43.2,8.8){\small $\mathbf{+}$}

        \put(1,9){\footnotesize Local maps}
        \put(17.5,4.5){\footnotesize Goal}
        \put(22.1,4.5){\begin{turn}{90} \footnotesize 2\end{turn}}
        \put(15.5,2.3){\footnotesize Velocity}
        \put(22.1,2.3){\begin{turn}{90} \footnotesize 2\end{turn}}

        \put(24.5,4.5){\begin{turn}{90} \footnotesize fully conn.\end{turn}}
        \put(26.5,4.5){\begin{turn}{90} \scriptsize ReLU\end{turn}}
        \put(32,4.5){\begin{turn}{90} \footnotesize tiling\end{turn}}
        \put(28.5,3){\begin{turn}{90} \footnotesize 64\end{turn}}
        \put(36.3,7.4){\scriptsize 64 vectors}

        \put(45.5,10){\begin{turn}{90} \scriptsize {3x3} \footnotesize{conv} \scriptsize {+ ReLU}\end{turn}}
        \put(55.5,10){\begin{turn}{90} \scriptsize {3x3} \footnotesize{conv} \scriptsize {+ ReLU}\end{turn}}
        \put(66,10){\begin{turn}{90} \scriptsize {3x3} \footnotesize{conv} \scriptsize {+ ReLU}\end{turn}}

        \put(76,10){\begin{turn}{90} \footnotesize{flatten}\end{turn}}
        \put(81,10){\begin{turn}{90} \footnotesize{fully conn.}\end{turn}}
        \put(80.5,8.1){\tiny {ReLU}}
        \put(86,10){\begin{turn}{90} \footnotesize{fully conn.}\end{turn}}
        \put(85.5,8.1){\tiny {ReLU}}
        \put(91,11.5){\begin{turn}{90} \footnotesize{dueling}\end{turn}}

        \put(49.5,12.7){\scriptsize 64 filters}
        \put(59.8,12.7){\scriptsize 64 filters}
        \put(70,12.7){\scriptsize 64 filters}

        \put(78.7,7.8){\begin{turn}{90} \footnotesize 4096\end{turn}}
        \put(83.7,8.5){\begin{turn}{90} \footnotesize 512\end{turn}}
        \put(88.7,8.5){\begin{turn}{90} \footnotesize 512\end{turn}}

        \put(93.7,6){\begin{turn}{90} \footnotesize 28\end{turn}}
        \put(93.7,11.7){\begin{turn}{90} \footnotesize 1\end{turn}}
        \put(98.7,8.7){\begin{turn}{90} \footnotesize 28\end{turn}}

        \end{overpic}
        \caption{The architecture of our CNN-based dueling DQN network. This network takes three local maps and 
        a vector with local goal and robot velocity as input and outputs the Q values of 28 discrete actions.}
        \label{net}
\end{figure*}
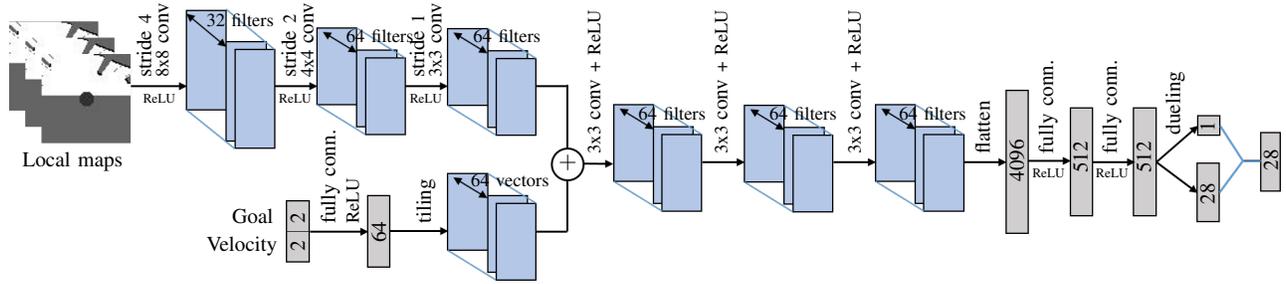

The reward function in reinforcement learning implicatly specifies what the agent is encourage to do.
Our objective is to avoid collisions during navigation and minimize the mean arrival time of the robot. 
A reward function is designed to guide robots to achieve this objective:
\begin{equation}
        r_{t} =  ({{r}^g})_{t} + ({{r}^c})_{t} + ({{r}^s})_{t}
\label{r1}
\end{equation}
The reward $r$ at time step $t$ is a weighted sum of three terms: ${{r}^g}$, ${{r}^c}$ and  ${{r}^s}$. 
In particular, the robot is 
awarded by $({{r}^g})_{t}$ for reaching its goal:
\begin{equation}
        ({{r}^g})_{t} = \left\{\begin{matrix*}[l]
                r_{arr} & \textrm{if} \left \| \textbf{p}_t - \textbf{g} \right \| < 0.2 \\  
                \epsilon(\left \| \textbf{p}_{t-1} - \textbf{g} \right \| - \left \| \textbf{p}_t - \textbf{g} \right \|)  & \textrm{otherwise}
                \end{matrix*}\right.
        \label{r2}
\end{equation}
When the robot collides with other obstacles in the environment, it is penalized by $({{r}^c})_{t}$:
\begin{equation}
        ({{r}^c})_{t} = \left\{\begin{matrix*}[l]
                r_{col} & \textrm{if collision}\\  
                0  & \textrm{otherwise}
                \end{matrix*}\right.
        \label{r3}
\end{equation}
And we also give robots a small fixed penalty ${{r}^s}$ at each step.
We set $r_{arr}$ = 500, $\epsilon$ = 10, $r_{col}$ = -500 and ${{r}^s} = -5$ in the training procedure.

\subsubsection{Network Architecture}
We define a CNN-based deep convolutional neural network that computes the action-value function for each actions. The input of the
network has three local maps $\textbf{M}$ with $60 \times 60$ grey pixels and a four-dimensional vector with local goal $\textbf{g}$ 
and robot velocity $\textbf{v}$. The output of the network is the Q-values for actions.
The architecture of our deep Q-value network is shown in Figure \ref{net}. 
The input costmaps are fed into a $8 \times 8$ convolution with stride 4, followed by a $4 \times 4$
convolution with stride 2 and a $3 \times 3$ convolution with stride 1. The local goal and robot velocity form a four-dimensional vector, which is
processed by one fully connected layer, and is then pointwise added to each point in the response map of processed costmap by tiling the output 
over the special dimensions. The result is then processed by three $3 \times 3$ convolutions and three fully connected layers with 512, 512 units respectively, 
and then fed into the dueling network architecture, 
after which the network outputs the Q values of 28 discrete actions.
\subsection{Curriculum Learning}
Curriculum learning \cite{bengio2009curriculum} is a learning strategy in machine learning, which starts from easy instances
and then gradually handles harder ones. In this work, we use Gazebo simulator \cite{koenig2004design} to build an environment with multiple obstacles. 
As the training progresses, we gradually increase the number of obstacles in the environment, and also the distance from the starting 
point of the target point gradually increases. This makes our strategy training from easy situation to difficult ones. At the same time, 
the position of each obstacle and the start and end points of the robot are automatically random during all training episodes. 
One training scene is shown in Fig. \ref{fig:scenarios:gazebo}.
\section{Experiments}
In this section, we present experiment setup and evaluation in both simulated and real environments. We quantitatively 
and contrastively evaluate the DQN-based navigation policy to show that it outperforms other related methods
in multiple indicators. Moreover, we also performed qualitative tests on real robots, and also integrated our obstacle 
avoidance policy into the navigation framework for long-range navigation testing.
\label{exp}
\subsection{Reinforcement Learning Setup}
The training experiments are conducted with a customized differential drive robot in a simulated virtual environment 
using Gazebo. A  180-degree laser scanner is mounted on the front of the robot as shown in 
Fig. \ref{fig:scenarios:gazebo}. The system parameters are empirically determined in terms of both the performance and our computation 
resource limits as listed in Table \ref{table1}.
\begin{table}[htbp]
        \centering
            \caption{System Parameters}
                \newcommand{\tabincell}[2]{\begin{tabular}[t]{@{}#1@{}}#2\end{tabular}}
                \begin{tabular}[t]{rl}
                        \toprule
                         Parameter& Value \\
                        \midrule
                        \tabincell{c}{learning rate}& \tabincell{c}{\textbf{$5 \times 10^{-4}$}} \\
                        \tabincell{c}{discount factor}& \tabincell{c}{\textbf{$0.99$}} \\
                        \tabincell{c}{replay buffer size}& \tabincell{c}{\textbf{$2 \times 10^{5}$}} \\
                        \tabincell{c}{minibatch size}& \tabincell{c}{\textbf{$1024$}} \\
                        \tabincell{c}{image size}&\tabincell{c}{\textbf{$60 \times 60$}} \\
                        \tabincell{c}{episode length}& \tabincell{c}{\textbf{$300$}} \\
                        \tabincell{c}{initial exploration}& \tabincell{c}{\textbf{$1$}} \\
                        \tabincell{c}{final exploration}& \tabincell{c}{\textbf{$0.1$}} \\
                        \bottomrule
                        \label{table1}
                \end{tabular}
\end{table}

The implementation of our algorithm is in TensorFlow, 
and we train the deep Q-network in terms of objective function Eq. \ref{ddqn} 
with the Adam optimizer \cite{kingma2014adam}. The training hardware is a computer with 
an i9-9900k CPU and a single NVIDIA GeForce 2080 Ti GPU. The entire training process (including exploration and training time) 
takes about 10 hours for the policy to converge to a robust performance.

\begin{figure}
        \centering
        \subfigure[] {\includegraphics[width = 0.48\linewidth]{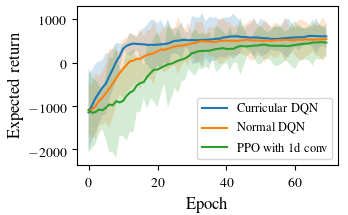}\label{fig:compare1}}
        \subfigure[] {\includegraphics[width = 0.48\linewidth]{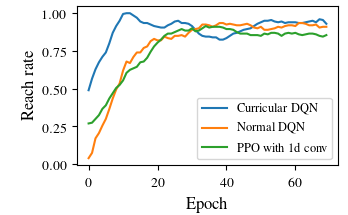}\label{fig:compare2}}
        \subfigure[] {\includegraphics[width = 0.48\linewidth]{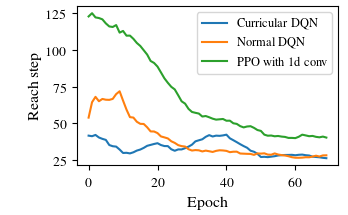}\label{fig:compare3}}
        \subfigure[] {\includegraphics[width = 0.48\linewidth]{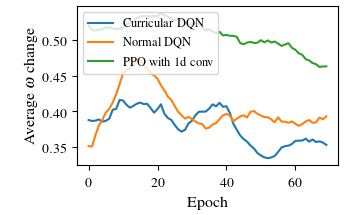}\label{fig:compare4}}
        \caption{Indicator curves of different methods during training.}
        \label{fig:compare}
\end{figure}
\subsection{Experiments on simulation scenarios}
\subsubsection{Performance metrics}
To compare the performance of our approach with other methods over various test cases, we define the following 
performance metrics.
\begin{itemize}
        \item \textit{Expected return $E_r$} is the average of the sum of rewards of episodes.
        \item \textit{Success rate $\bar{\pi }$} is the ratio of the episodes in which the robot reaching the goals within a certain step without 
        any collisions over the total episodes.
        \item \textit{Reach step $\bar{s}$}  is the average number of steps required to successfully reach the target point without 
        any collisions.
        \item \textit{Average angular velocity change $\triangledown \omega $} is the average of the angular velocity changes for each step, which reflects the 
        smoothness of the trajectory.
\end{itemize}
\subsubsection{Comparative experiments}
\begin{figure}
        \centering
        \subfigure{\includegraphics[width = 0.48\linewidth]{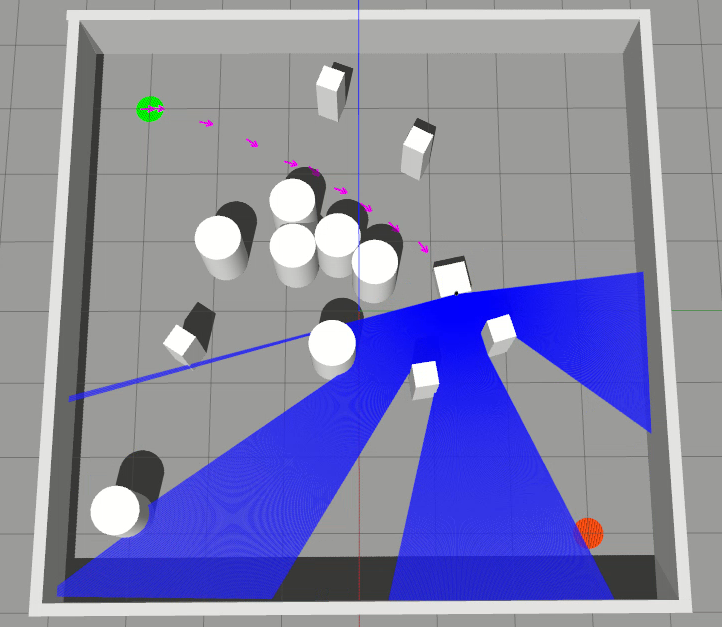}}
        \subfigure{\includegraphics[width = 0.48\linewidth]{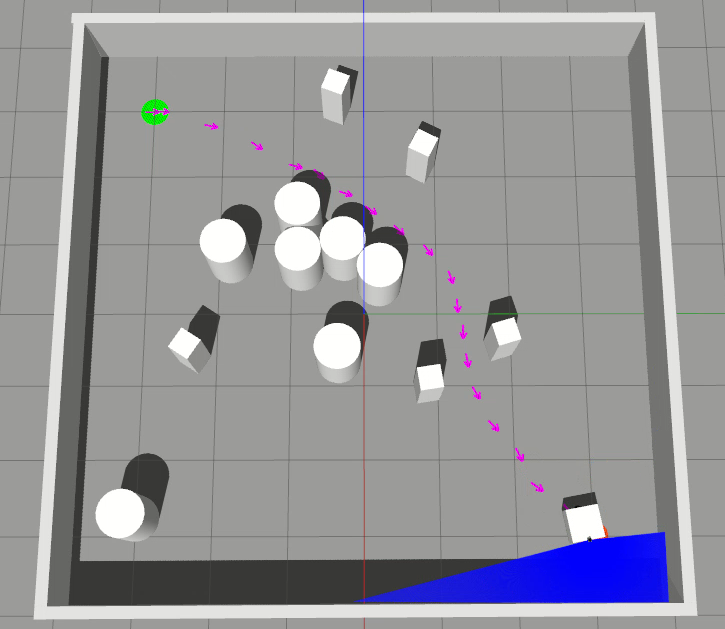}}
        \caption{ A test case of our curricular DQN policy in a test scenario, the green dot represents the starting point, 
        the red dot represents the end point, and the robot's trajectory is marked with purple arrows.}
        \label{fig:example}
\end{figure}

\begin{table}[htbp]
        \centering
            \caption{Indicators values of various methods}
                \newcommand{\tabincell}[2]{\begin{tabular}[t]{@{}#1@{}}#2\end{tabular}}
                \begin{tabular}[t]{rcccc}
                        \toprule
                        Method & $E_r$ & $\bar{\pi }$ & $\bar{s}$ & $\triangledown \omega $ \\
                        \midrule
                        \tabincell{c}{PPO with 1d conv}&\tabincell{c}{467.87}&\tabincell{c}{0.85} & \tabincell{c}{40.19}  & \tabincell{c}{0.46} \\
                        \specialrule{0em}{1pt}{1pt}
                        \tabincell{c}{Normal DQN}&\tabincell{c}{547.43}&\tabincell{c}{0.91} & \tabincell{c}{27.76}  & \tabincell{c}{0.39} \\
                        \specialrule{0em}{1pt}{1pt}
                        \tabincell{c}{Curricular DQN}&\tabincell{c}{\textbf{617.04}}&\tabincell{c}{\textbf{0.94}} & \tabincell{c}{\textbf{26.13}}  & \tabincell{c}{\textbf{0.35}} \\
                        \bottomrule
                        \label{table2}
                \end{tabular}
\end{table}
We compare our curricular DQN policy with normal (non-curricular) DQN policy and PPO with one-dimensional convolution network 
\cite{long2018towards} in our tests. As shown in Fig. \ref{fig:compare}, our DQN-based policy has significant 
improvement over PPO policy in terms of expected retrun, success rate, reach step and average angular velocity change, and curricular DQN 
policy has
also a slight improvement on multiple indicators. In the tests with more obstacles, the specific indicators values of various method 
are shown in the Table \ref{table2}. Fig. \ref{fig:example} shows a test case of our curricular DQN policy in a test scenario.

\begin{figure}
        \centering
        \subfigure{\includegraphics[width = 0.9\linewidth]{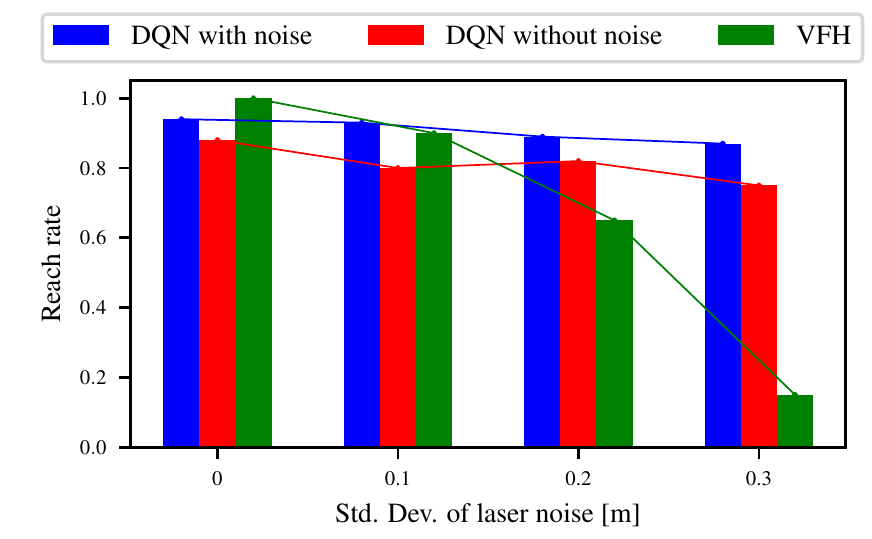}}
        \caption{ Reach rate over laser noise. DQN policy trained with sensor noise compared to 
        policy without sensor noise and traditional VFH method.}
        \label{fig:noise}
\end{figure}
\subsubsection{Robustness to noise}
Fig. \ref{fig:noise} depicts that the performances of our DQN-based policy and the traditional vector field histogram (VFH) method \cite{ulrich2000vfh}
vary with the noise error of the laser sensor data in an environment with many obstacles. Results show that 
our DQN-based policy is resilient to noise, and laser noise heavily influences the reach rate of VFH. This is expected since 
VFH uses obstacle clearance to calculate its objective function, and such a greedy approach often guides the robot to local minima. 
More importantly, the learned policy (the DQN policy with noise in the Fig. \ref{fig:noise}) will work better when using the same noise 
variance as the test environment during training.
\subsection{Navigation in real-world}
\begin{figure}
        \centering
        \begin{overpic}[width=0.9\linewidth]{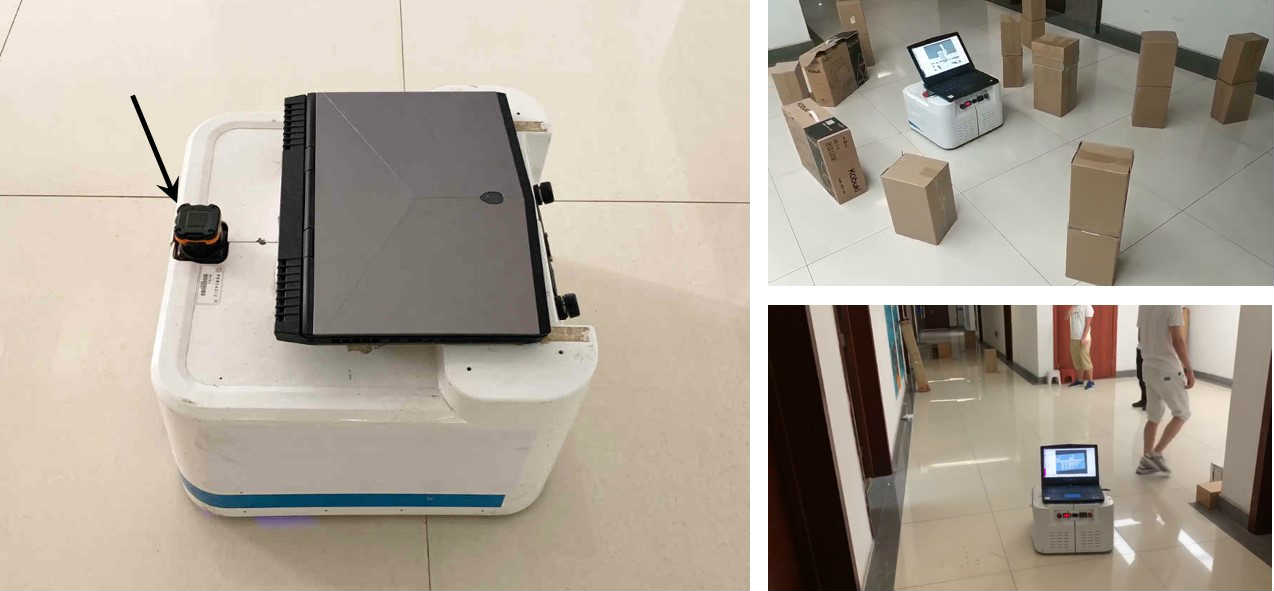}
        \put(7,41){\fontsize{7pt}{\baselineskip}\selectfont 2d laser scanner}
        \end{overpic}
        \caption{Our robot chassis with a laptop and a Hokuyo UTM-30LX laser scanner (left). Real test environment(right), including 
        difficult obstacle environments (upper right) and long-range corridor test (bottom right).}
        \label{fig:kejia}
\end{figure}

To further verify the generalization and effectiveness of our learning policy, we use our robot chassis to do experiments in  
real-world. As shown in Fig. \ref{fig:kejia}, our robot platform is a differential wheel robot with a Hokuyo UTM-30LX scanning laser rangefinder and a laptop with 
an i7-8750H CPU and a NVIDIA 1060 GPU. The robot pose and velocity are provided by a particle filter based state estimator. An 
occupancy map is constructed from laser measurement, from which an egocentric local map is cropped to fixed size $6.0 \times 6.0$m 
and resolution 0.1m at each cycle. 

We used paper boxes to build different difficult environments for testing. As shown in Fig. \ref{fig:kejia}, when the robot 
confronts obstacles, the trained policy succeeds in providing a reactive action command that drives the robot away from obstacles. In the 
long-range experiments, our robots navigate safely in corridors with obstacles and pedestrians. A video for real and simulated 
navigation experiments can be found at \url{https://youtu.be/Eq4AjsFH_cU}.
\section{CONCLUSIONS}
\label{conclusion}
In this paper, we propose a model-free deep reinforcement learning algorithm to
improve the performance of autonomous decision making in
complex environments, which directly maps egocentric local occupancy maps to an agent’s steering commands in
terms of target position and movement velocity. 
Our approach is mainly based on  dueling double DQN with prioritized experience reply, and integrate curriculum learning techniques 
to further enhance our performance. Finally, both qualitative and quantitative results show
that the map-based motion planner outperforms other related
DRL-based methods in multiple indicators in simulation environments and is easy to be deployed to a robotic platform.

\addtolength{\textheight}{-12cm}   

\bibliographystyle{IEEEtran}
\bibliography{IEEEabrv,my}
\end{document}